\documentclass[letterpaper, 10pt, conference]{ieeeconf}

\IEEEoverridecommandlockouts                              % This command is only needed if 
\usepackage[english]{babel}

\usepackage[letterpaper,top=2cm,bottom=2cm,left=3cm,right=3cm,marginparwidth=1.75cm]{geometry}

\usepackage{amsmath}
\usepackage{graphicx}
\usepackage{subcaption}
\usepackage[colorlinks=true, allcolors=blue]{hyperref}
\usepackage{subfiles}
\usepackage{xcolor}
\usepackage{todonotes}
\usepackage{booktabs}
\usepackage{colortbl}
\let\labelindent\relax
\usepackage{enumitem}
\definecolor{llm_orange}{RGB}{227, 140, 71}
\definecolor{llm_blue}{RGB}{105, 156, 219}
\hypersetup{
    colorlinks=true,
    linkcolor=llm_blue,
    citecolor=llm_blue,
    urlcolor=llm_blue
}

\title{\LARGE \bf
AGENTS-LLM: \underline{A}ugmentative \underline{GEN}eration of Challenging \underline{T}raffic \underline{S}cenarios with an Agentic LLM Framework}

\author{%
Yu Yao$^{1\dagger*}$, %
Salil Bhatnagar$^{2\ddagger*}$, %
Markus Mazzola$^{1}$, %
Vasileios Belagiannis$^{2}$, \\%
Igor Gilitschenski$^{3,6}$, %
Luigi Palmieri$^{1}$, %
Simon Razniewski$^{4}$ and % 
Marcel Hallgarten$^{1,3,5}$%
\vspace{0.2cm}
\\
$^{1}$Robert Bosch GmbH \quad
$^{2}$Friedrich-Alexander-Universität Erlangen-Nürnberg \quad
$^{3}$University of Toronto \quad \\
$^{4}$3ScaDS.AI \& TU Dresden \quad
$^{5}$University of T{\"u}bingen \quad
$^{6}$Vector Institute \quad \\
\vspace{-0.1cm}
{\tt\small \href{https://github.com/mh0797/Agents-LLM/}{{\color{llm_orange}{https://github.com/mh0797/Agents-LLM/}}}}
\thanks{$^{\dagger}$Corresponding author}%
\thanks{$^{\ddagger}$Work performed while with Robert Bosch GmbH}%
\thanks{$^{*}$Authors contributed equally}%
}

\begin{document}
\maketitle
\thispagestyle{empty}
\pagestyle{empty}

\begin{abstract}
Rare, yet critical, scenarios pose a significant challenge in testing and evaluating autonomous driving planners.
Relying solely on real-world driving scenes requires collecting massive datasets to capture these scenarios.
While automatic generation of traffic scenarios appears promising, data-driven models require extensive training data and often lack fine-grained control over the output.
Moreover, generating novel scenarios from scratch can introduce a distributional shift from the original training scenes which undermines the validity of evaluations especially for learning-based planners.
To sidestep this, recent work proposes to generate challenging scenarios by augmenting original scenarios from the test set. However, this involves the manual augmentation of scenarios by domain experts. An approach that is unable to meet the demands for scale in the evaluation of self-driving systems.
Therefore, this paper introduces a novel LLM-agent based framework for augmenting real-world traffic scenarios using natural language descriptions, addressing the limitations of existing methods.
A key innovation is the use of an agentic design, enabling fine-grained control over the output and maintaining high performance even with smaller, cost-effective LLMs.
Extensive human expert evaluation demonstrates our framework's ability to  accurately  adhere to user intent, generating high quality augmented scenarios comparable to those created manually.
\end{abstract}

\section{Introduction}

Generalization to rare and critical scenarios such as dangerous and erratic driving \cite{wiederer2022anomaly}, is paramount to safety in autonomous driving (AD). Its validation demands large-scale testing across diverse datasets.
Real-world datasets~\cite{nuplan,sun2020scalability} offer the advantage of testing planners with the same distribution of traffic scenes encountered during deployment.
However, they underrepresent challenging and safety-critical scenarios, commonly referred to as long-tail events.

The rarity of long-tail events makes them prohibitively expensive to collect while their safety-critical nature raises ethical concerns about targeted collection.
Thus, recent methods leverage generative models to synthesize test scenarios and edge cases~\cite{zhao2024,Chitta2024ECCV,tian2024}.
However, they often lack fine-grained control over the output and generating novel scenarios \cite{bergamini2021simnet,tan2021scenegen} from scratch can introduce a distributional shift from the original training scenes which undermines the validity of evaluations especially for learning-based planners.
In contrast, augmentative scene generation makes small changes to existing scenes to create novel, yet realistic, scenarios representing critical corner cases.
Recent work~\cite{Hallgarten2024interPlan} relies on domain experts to augment recorded scenes to create challenging test cases based on real-world data.
While this approach has proven highly effective~\cite{Hallgarten2024interPlan}, its manual nature limits scalability.

\begin{figure*}[t!]
    \centering
    \includegraphics[width=0.9\textwidth, angle=0]{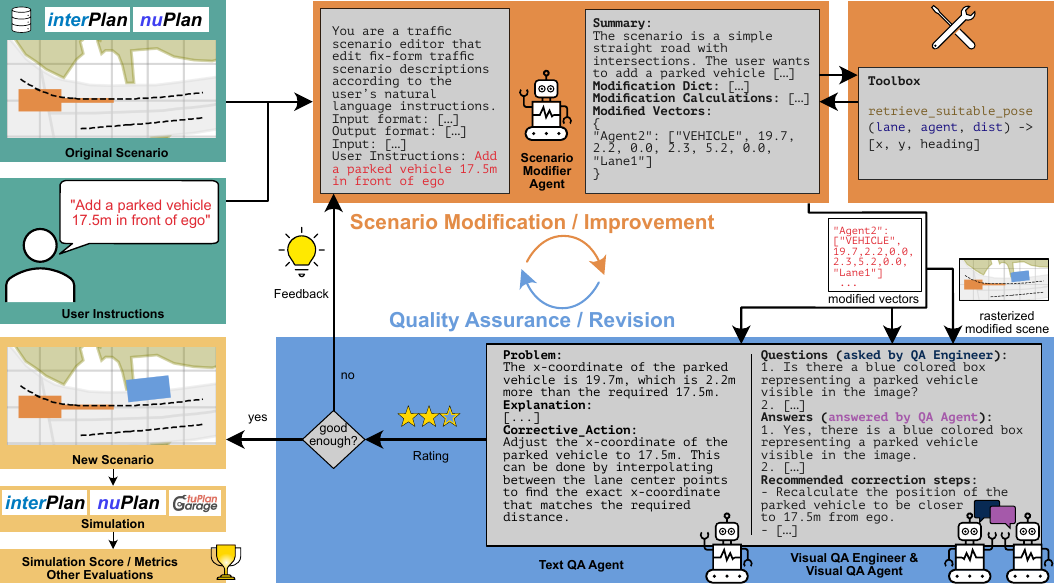}
    \caption{Scenario modification framework. The Scenario Modifier Agent generates a modified scenario based on an original scenario and user instructions. To do this, the modifier agent has the option to make calls to an external function. An optional Quality Assurance loop can be used to evaluate the result and request corrections from the modification agent. Two alternatives exist for the QA loop: text and visual QA.}
    \label{fig:framework}
\vspace{-0.5cm}
\end{figure*}

In this work, we automate this augmentation process by proposing a Large Language Model (LLM)-assisted framework for natural language-guided augmentation of real-world traffic scenarios (Figure~\ref{fig:framework}).
While LLM-assisted methods to generate traffic scenes from scratch exist, ours is the first to augment real-world scenarios guided by high-level natural language descriptions of the intended modifications.
Furthermore, to the best of our knowledge, we are the first to leverage an agentic design pattern~\cite{react} for this task.
This provides fine-grained control over the output—a feature lacking in many current generative methods—and allows us to use smaller, cheaper LLMs to achieve performance comparable to large, expensive models.
Additionally, we pioneer the quantitative evaluation of augmented scenarios using pairwise comparison by human domain experts and the Elo~\cite{chiang2024chatbot} rating system.

Our contributions are as follows:
\begin{enumerate}[wide, labelwidth=!, labelindent=0pt, label=\textbf{\arabic*})]
    \item We introduce an LLM agent-based framework for modifying traffic scenarios using natural language and examine its performance under various LLMs.
    \item We show that our agentic framework enables cost-effective and compact LLMs to achieve performance comparable to large, expensive frontier models.
    \item Finally, we show that our framework generates scenarios which challenge SotA planning algorithms.
\end{enumerate}

\section{Related Work}
Scenario generation for the testing and evaluation of AD systems has been extensively studied \cite{cai2022, ding2023surveysafetycriticaldrivingscenario,scenecontrol2024,queiroz2019geoscenario} due to the safety-critical nature of the topic.
Aside from data-driven methods \cite{li2022}, several holistic approaches \cite{zipfl2023comprehensive, ma2021} have also been introduced in the past. 

\paragraph{Language-based Scenario Generation}
Text-to-Drive~\cite{nguyen2024} introduced a method for generating driving behavior using LLMs.
While behavior modeling is an important topic, the complexity involved in faithfully capturing different driving styles requires the use of dedicated models.
Therefore, our work focuses on the generation of the traffic scene itself.

A second line of research deals with language guided manipulation of Scenic~\cite{fremont2019scenic} code for use within the CARLA simulator.
Chatscene~\cite{zhang2024chatscene} focused on the creation of safety-critical scenes,
while Miceli-Barone et al.~\cite{micelibarone2023} developed an LLM \emph{assistant} designed for interactive, turn-based generation of Scenic code.
In contrast, our work is not limited to a specific simulator like CARLA. Instead, we use a generic text-based scenario description, which can be easily adapted to different formats. This is demonstrated by importing our scenarios into nuPlan~\cite{nuplan}, which provides access to many state-of-the-art planner algorithms.

Finally, LCTGen~\cite{tan2023language} represents a hybrid approach where a natural language prompt is processed by  an LLM into an abstract scene representation, referred to as \emph{code}. A trained neural network then generates the actual scene from this code. While this approach requires training a separate model in addition to the LLM, our framework does not require training nor fine-tuning of the LLMs involved.

\paragraph{Data-driven Scenario Generation}
In a parallel line of research, data-driven models have emerged as a powerful tool for generating detailed and realistic scenarios.
These include approaches where a diffusion model generates a bird's eye view (BEV) image, followed by a data-driven \cite{Pronovost2023, Chitta2024ECCV} or heuristic \cite{sun2024drivescenegen} component to extract vectorized representations for map, scenario and behavior.
Alternatively, diffusion processes can also operate directly in the space of vectorized scenario representations~\cite{scenecontrol2024}.
Instead of sampling from the distribution of training datasets, RealGen~\cite{ding2024realgen} proposes to train a model to combine existing scenarios retrieved from a database into novel scenarios.
By following an agentic design pattern using LLM-based agents, we sidestep the requirement for large training datasets that comes with data-driven methods.
At the same time, we take advantage of the instruction-following capability built into LLMs.

\paragraph{Augmentative Scenario Generation}
Rather than generating scenarios from scratch, real-world scenarios can be augmented with additional actors, objects, and obstacles to make them more challenging~\cite{yang2023unisim, zhang2022adversarial, hallgarten2024stay}.
Using real-world driving data as a basis ensures minimal distributional shift for data-driven planners~\cite{bahari2022vehicle}.
Previously, \emph{interPlan}~\cite{Hallgarten2024interPlan} proposed to use such augmentations for generating challenging scenarios across five categories: Passing construction zones, encountering an accident site, avoiding jaywalkers, nudging around parked vehicles, and overtaking obstacles with oncoming traffic.
However, all scenarios are manually generated based on experts' knowledge. 
In this paper, we leverage LLMs in an agentic framework to automatically generate scenario augmentations from a natural language description of the intended modification

\section{Method}
We present a scenario augmentation framework capable of generating realistic and challenging driving scenarios that can be executed in a closed-loop simulation.
The framework, shown in \autoref{fig:framework} is based on an agentic design and is compatible with interPlan's scenario augmentation interface.
Below, we provide a basic introduction on agentic LLM-agents, before describing our modification framework.

\subsection{Agentic Framework}
In the context of large language models, \textit{agentic frameworks}, also called LLM agents, refer to design approaches that give models the ability to perform tasks in ways that resemble autonomous agents~\cite{react,wu2024autogen}.
Unlike standard LLM usage, where the response to an input is generated through a single call to the LLM, agentic frameworks involve layered or repeated calls to LLMs.
This extends the \textit{chain-of-thought} concept~\cite{chain-of-thought}, where LLM responses improve due to allowing the model to carry out multiple explicit reasoning steps which increases the per-task computation budget.
Agentic frameworks provide LLMs with structures for planning, reasoning, function calling \cite{schick2023toolformer}, and adapting based on intermediate outcomes, making them potentially more versatile and effective in scenarios requiring sustained, context-aware decision-making.

\subsection{Scenario Modification Framework}
\label{sec:framework}
\autoref{fig:framework} shows a high-level overview of the proposed framework, which comprises a number of LLMs taking the roles of interacting agents.
Initially, a \emph{Scenario Modifier Agent} (SMA) receives the original scenario together with a set of user instructions and generates an updated scenario containing the requested modifications.
Advanced prompting techniques are employed to allow the SMA to understand the initial scenario and then generate modified traffic agents or objects that align with the user instructions.
Specifically, we explore the option of allowing LLMs to make function calls in order to retrieve relevant coordinates along lanes.
The output of the SMA is either directly processed by interPlan or passed through a quality assurance (QA) loop consisting of one or more \emph{Quality Assurance Agents}, who's goal it is to verify if the SMA's output aligns with the user intention. 
In this paper, we explore two different QA strategies: text-only and hybrid visual-text.
\subsubsection{Text QA Agent} In the text-only variant, a QA agent receives the initial scenario representation, the user instructions, the modified traffic agent vectors generated by the SMA and a list of common problems compiled from typical mistakes observed during initial experimentation.
Given this input, the QA agent summarizes the initial scenario and user intent, plans verification questions that help evaluate the SMA's output and answers these questions and finally, rate the SMA's output in three categories: Compliance with User Instructions, Realism, and Logical Consistency.
In each category, a rating from 1 to 5 is generated and if the average rating is less than 4, the QA agent generates step-by-step feedback by identifying the problem, explaining the reason behind the error and suggesting corrective actions.
This feedback is sent back to the SMA, which regenerates the scenario.
\subsubsection{Visual QA Agent and Engineer} In contrast, visual QA is a multi-modal, multi-agent approach, where an LLM, the \emph{QA Engineer} generates critical questions which help evaluate the SMA's output.
Next, a vision language model, the \emph{QA Agent}, receives a rendered BEV image of the modified scenario and the questions from the QA Engineer, with the task to answer these questions and retrieve relevant information about the modified scenario from the image.
Finally, the QA Engineer utilizes the output from the QA Agent to identify mistakes in the modification work and provide feedback to the SMA.

\definecolor{user_input_green}{RGB}{89,168,156}
\definecolor{scenario_modifier_orange}{RGB}{227 140 71}
\definecolor{prompt_backround_grey}{RGB}{206 206 206}
\definecolor{darkblue}{RGB}{0 0 139}
\begin{figure*}[t]
    \centering
    \fcolorbox{user_input_green}{user_input_green}{%
    \fcolorbox{black}{prompt_backround_grey}{\scriptsize
    % \begin{minipage}{.45\textwidth}
    \begin{minipage}[t][5.5cm][t]{.45\textwidth}
    \begin{center}
        \textbf{Input prompt:}
    \end{center}
    \vspace{-0.2cm}
    % \par
    \texttt{You are a traffic scenario editor that edits fix-form traffic scenario descriptions according to the user's natural language instructions.}
    \par
    \texttt{\textbf{Input format:}}
    % \par
    \texttt{Input vectors \{"Agent":} [...]
    \par
    \texttt{\textbf{Instruction format:} natural language} [...]
    \par
    \texttt{\textbf{Output format:}}
    \par
    \texttt{Insights: Take your time and step by step } [...]
    \par
    \texttt{\textbf{Input:}}
    \par
    \texttt{\{"Agent1": ["EGO\_VEHICLE",} [...]
    \par
    \texttt{\{"Lane1": [1, "Eastwards",} [...]
    \par
    \texttt{\textbf{User Instructions:}}
    \par
    {\color{red}\texttt{add a parked vehicle in front of/in the travel direction of ego at an approx. distance 21.4m away from ego. Assume a slight offset (anything randomly between 0 to 1.5m) from lane center points as it is parked slightly towards the left lane boundary.} [...]}
    \par
    \texttt{\textbf{Output:}}
    \end{minipage}
    }%
    }%
    \vspace{.05\textwidth}
    \fcolorbox{scenario_modifier_orange}{scenario_modifier_orange}{%
    \fcolorbox{black}{prompt_backround_grey}{\scriptsize
    % \begin{minipage}{.45\textwidth}
    \begin{minipage}[t][5.5cm][t]{.45\textwidth}
    \begin{center}
        \textbf{Response:}
    \end{center}
    \vspace{-0.2cm}
    % \par
    \texttt{\textbf{Insights:}}
    \par
    \texttt{The scenario consists of a single lane road with ego vehicle traveling eastwards} [...]
    \par
    \texttt{\textbf{Summary:}}
    \par
    \texttt{The user wants to add a parked vehicle in front of ego at a distance of 21.4m, slightly offset from the lane center towards the left lane boundary} [...]
    \par
    \texttt{\textbf{Modification Dict:}}
    \par
    \texttt{\{"Action": "add", "Modified\_Agent": "Agent2", } [...]
    \par
    \texttt{\textbf{Modification Calculations:}}
    \par
    \texttt{Step 1: Calculate the approximate position of the parked vehicle.} [...]
    \par
    \texttt{Step 2: Retrieve the closest lane anchor at the approximate position of} [...]
    \par
    \texttt{\textbf{Modified Vectors:}}
    \par
    {\color{red}\texttt{\{"Agent2": ["VEHICLE", 21.4, 2.6,} [...]}
    \end{minipage}
    }%
    }%
    \vspace{-2em} % hack to remove excessive space between text boxes and caption
    \caption{Exemplary input prompt and response from our framework.}
    \label{fig:prompt}
\vspace{-0.5cm}
\end{figure*}

\subsection{Scenario Representation and Tool Use}
\label{sec:tool_use}
An example input prompt for the SMA and the corresponding response are shown in \autoref{fig:prompt}.
To allow LLMs to process traffic scenarios, we use a text-based scenario representation consisting of separate lists for traffic agents, lanes/lane connectors and areas.
Each list element corresponds to an entity of the respective category represented by a vector of attributes.
An overview is given in \autoref{tab:input_vectors}.
\begin{table}[]
    \centering
    \resizebox{\columnwidth}{!}{
    \begin{tabular}{p{1.0cm}p{5.0cm}}
        \toprule
        \textbf{Entity} & \textbf{Input Vectors} \\
        \midrule
         Agent & Agent type, center coordinates, heading, width, length, velocity, lane ID \\
         \rowcolor{llm_blue!20}
         Lane & Lane ID, travel direction, relative direction to ego, width, speed limit, lane coordinates \\
         Lane \mbox{Connector} & From lane, to lane, traffic light state, turn type, speed limit, lane coordinates \\
         \rowcolor{llm_blue!20}
         Area & Boundary points \\
        \bottomrule
    \end{tabular}
    }
    \caption{Input vectors for different entities.}
    \label{tab:input_vectors}
    \vspace{-0.5cm}
\end{table}

Importantly, we explore two alternatives for representing lanes:
a list of centerline coordinates sampled at a distance of 5 meters (polyline format)
or the four control points of a cubic Bézier curve (Bézier format).
In order to use the Bézier format, we include instructions in the prompt of the SMA on how to \textit{call a function} to retrieve a point of interest along a lane or lane connector given the control points of the Bézier curve.
During execution, the LLM representing the SMA needs to decide when to call this function and what arguments to pass.

\subsection{Dataset and Simulation Framework}
In this work, we use the interPlan~\cite{Hallgarten2024interPlan} scenarios as ground-truth for development and later assess the quality of our automatically generated scenarios by presenting both to human experts, who are asked to rank them.
Thus, we employ our framework to generate modification vectors for interPlan's scenario augmentation interface.
Further, we use the nuPlan~\cite{nuplan} simulator to run the generated scenarios in closed loop.
Due to its wide adoption in planning, nuPlan offers interoperability with many state-of-the-art planning algorithms~\cite{scheel2022_urban_driver, huang2023_gameformer, hallgarten2023prediction, huang2024dtpp, Dauner2023CORL, chekroun2024mbappe}.
We demonstrate that our generated scenarios are able to challenge state-of-the-art planners in closed-loop simulation.

\section{Experiments and Evaluation}

In order to assess the ability to faithfully follow user instructions and generate useful scenario modifications, we use our framework to recreate the 50 human-augmented scenarios from interPlan.
In doing so, we explore the design space in two different dimensions: the prompting strategy used in the agentic framework and the LLMs representing the agents.
For the latter, we cover three model classes: 1) frontier models, i.e., the best currently available LLMs, 2) utility models, i.e., commercial models which are less performant but more cost effective and 3) open-weight LLMs.
We choose GPT-4o, Gemini-1.5-Flash and Llama3.1-70B to represent each of the three classes, respectively.
All models are used in their pretrained form without any problem-specific fine-tuning.

In terms of prompting strategy, we examined a number of variants listed in \autoref{tab:variants}.
These differ by which of the components from \autoref{fig:framework} they include.
The baseline variant called "one-time-modifier" (\texttt{OTM}) consists of only the scenario modifier agent (SMA) prompted with the polyline lane format.
Based on \texttt{OTM}, we explore "function calling" (\texttt{FC}) by switching to the Bézier format and prompting the LLM-agent for tool use.
Orthogonally, we explore QA variants by activating either the text-only (\texttt{tQA}) or the visual (\texttt{vQA}) variant.

\begin{table}
    \centering
    \begin{tabular}{ll l}
        \toprule
        \textbf{Variant} & \textbf{Lane Rep.} & \textbf{QA Loop} \\
        \midrule
        one-time-modifier (\texttt{OTM}) & polyline & none \\
        function calling (\texttt{FC}) & Bézier & none \\
        text QA (\texttt{tQA}) & polyline & text-only \\
        visual QA (\texttt{vQA}) & polyline & visual \\
        \bottomrule
    \end{tabular}
    \caption{Prompting Strategies.}
    \label{tab:variants}
    \vspace{-0.5cm}
\end{table}

Assessing the quality of a scenario augmentation is difficult due to the lack of established metrics.
In this work, we focus on two aspects for quality assessment: \textit{placement accuracy} and \textit{visual appearance}.
Placement accuracy denotes the precision with which the framework is able to place traffic agents based on user instructions.
This is important for the ability to create traffic scenarios representing specific safety-critical situations.
For example, if the user intents to insert a stopped vehicle in front of an intersection such that it blocks the view, a few meters of error might lead to the vehicle actually being placed behind the intersection.
To measure placement accuracy, we match each LLM-modified traffic agent to the closest human-modified agent from interPlan via the Hungarian algorithm~\cite{hungarian}, and compute the displacement error as the distance between their center points in meters.

While a high placement accuracy is required for creating very specific scenarios, a deviation from the manual placement from interPlan does not generally equate to a misalignment with user intent.
For example, when generating a construction zone, the exact position and number of cones is typically less important than the overall location and extend of the construction area.
Therefore, to complement the displacement error metric, we propose to also assess the visual appearance of the modified scenario as a whole.
Inspired by Chatbot Arena \cite{chiang2024chatbot}, we tackle this problem using pair-wise ranking by human domain experts.
We presented experts from the autonomous driving research community with a side-by-side comparison of BEV images of the same scenario generated by two different models based on the same user instructions.
The experts rated which of the two model outputs they preferred, or if both were perceived as equal.
The order of match-ups were randomized and the identities of the models were hidden from the judges during rating.
While collecting these expert ratings is very time consuming, we believe that the value of human-based evaluation justifies the effort.

\section{Results}

\subsection{One-Time-Modifier Variant}
\begin{figure}[t]
    \centering
    \includegraphics[width=3in]{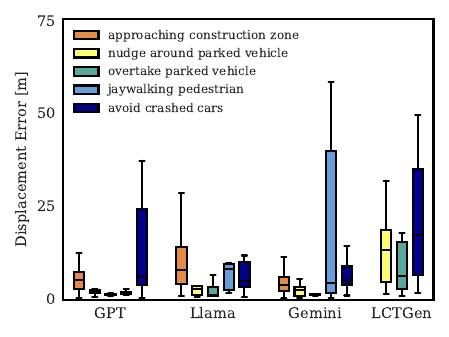}
    \caption{Displacement error by scenario type.}
    \label{fig:displacement_type}
    \vspace{-0.5cm}
\end{figure}
We begin by assessing the \texttt{OTM} variant of our framework.
\autoref{fig:displacement_type} shows the displacement error grouped by scenario type for the \texttt{OTM} variants with different LLMs as SMA.
We observe that the error is characterized by large outliers and differs significantly between scenario types.
This indicates that some types (i.e., "accident site" and "construction zone") are inherently more ambiguous than others.
In addition, different LLMs excel in different categories, although overall, the frontier model (GPT-4o) performs best.

In addition, we also included  LCTGen~\cite{tan2023language}, a strong recent language-based scenario modification baseline, which we adapted to the interPlan scenario catalogue.
However, LCTGen failed to generate any traffic agents for the "jaywalker" and "construction site" types, due to its vehicle-centric design.
For the scenario types where LCTGen successfully generated traffic agents, the error is significantly larger on a per category basis.
We speculate that this is due to the intermediate representation used by LCTGen, which quantizes the desired vehicle positions into discrete range and heading brackets.
This cuts off the generator component from the detailed information in the user instructions that would be necessary for accurate placement.

In order to gain an understanding for the cause behind the observed errors, we render each scenario generated using our framework as a BEV image.
We then manually evaluate each scenario to identify common failure cases.
Based on this analysis, we group errors into three categories:
\begin{enumerate}[wide, labelwidth=!, labelindent=0pt, label=\textbf{\arabic*})]
    \item \emph{Position Error}: The traffic agent is positioned at the wrong distance or completely offroad. This is most often due to a failure by the SMA to retrieve the correct lane anchor.
    \item \emph{Heading Error}: The heading of the traffic agent is wrong, while the position is approximately correct. This is typically due to the SMA ignoring the lane heading.
    \item \emph{Logic Error}: The SMA made an error during reasoning or calculation, e.g., by placing two vehicles on top of each other in the accident site scenario.
\end{enumerate}

\autoref{tab:manual_eval} lists the error count per category for the three LLMs.
Unsurprisingly, GPT-4o makes significantly fewer errors than the two smaller models.
We observe that "Position Error" is the dominant error type overall and Llama the model most affected by it.
In contrast, GPT-4o achieves both a low error count and a low displacement error in most scenario types, establishing it as a solid baseline for further analysis.
In conclusion, there is a significant performance gap between frontier LLMs and smaller models when using \texttt{OTM}, which is a relatively simple prompting strategy.

\begin{figure}[t]
    \centering
    \includegraphics[width=3in]{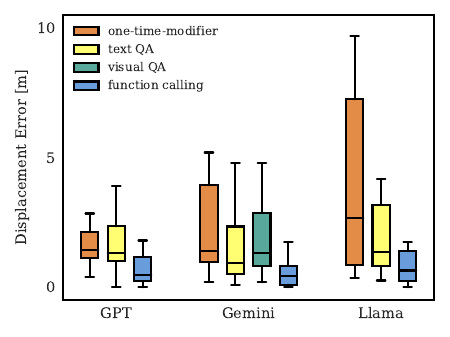}
    \caption{Displacement error by variant.}
    \label{fig:displacement_variant}
    \vspace{-0.5cm}
\end{figure}

\begin{table}
    \centering
    \resizebox{\columnwidth}{!}{
    \begin{tabular}{l>{\columncolor{llm_blue!20}}rrrr}
    \toprule
    \textbf{Model} &  \textbf{All}~$\downarrow$ & \textbf{Position}~$\downarrow$ & \textbf{Heading}~$\downarrow$ & \textbf{Logic}~$\downarrow$ \\
    \midrule
    GPT-4o \texttt{OTM} & 5 & 3 & 0 & 2\\
    Gemini-1.5-Flash \texttt{OTM} & 15 & 6 & 2 & 7\\
    Llama3.1-70B \texttt{OTM} & 16 & 11 & 3 & 2\\
    \bottomrule
    \end{tabular}
    }
    \caption{Error Count per Category.}
    \label{tab:manual_eval}
    \vspace{-0.5cm}
\end{table}

\subsection{Advanced Prompting Strategies}
Having established GPT-4o \texttt{OTM} as a solid baseline, we turn our attention to function calling (\texttt{FC}), text QA (\texttt{tQA}) and visual QA (\texttt{vQA}).
Our goal is to investigate, if it is possible to close the gap between frontier models and utility/open-weights models by leveraging these advanced prompting strategies.
Ultimately, this would help to circumvent the high costs associated with frontier models and reduce the reliance on closed-source commercial APIs.

\autoref{fig:displacement_variant} shows the displacement error for all scenarios of the types "jaywalker", "nudge around parked vehicle" and "overtake parked vehicle" plotted against different prompting strategies.
Note that we excluded "accident site" and "construction zone" from this comparison in order to reduce the influence of the ambiguity inherent to these scenario types.
For GPT-4o, we observe little improvement beyond the performance of \texttt{OTM}, except for a slight error reduction for \texttt{FC}.
However, for the other LLMs, there is a noticeable improvement with more advanced prompting techniques, with \texttt{FC} being the most effective variant and Llama the model which sees the clearest improvement.
This confirms our hypothesis that smaller models can achieve competitive placement accuracy using \texttt{FC}, which mitigates retrieval errors--the most common error source.

Turning our attention to the QA variants, we observe that surprisingly, \texttt{vQA} does not improve displacement error over \texttt{tQA} on average.
Note however that, due to the cost involved, \texttt{vQA} was only evaluated for Gemini.
Since \texttt{vQA} is a multi-agent design and, in addition, requires vision input, it involves processing a large number of tokens.
This makes it very cost inefficient when used with a frontier model such as GPT-4o, which already performs very well in the \texttt{OTM} variant.
In addition, the requirement for vision input also rules out Llama3.1.

\subsection{Human Expert Ranking}
The third pillar of our evaluation is an expert ranking conducted among eight variants of our framework and interPlan.
Overall, 5760 pairwise comparisons from nine experts from the autonomous driving research community were collected.
Based on this data, we computed Elo model strength and also 95\% confidence intervals via bootstrapping \cite{chiang2024chatbot}, which are shown in \autoref{tab:elo}.
Elo assigns a numerical rating to contestants based on their performance in head-to-head matches, with the rating difference between two players determining the expected outcome.
For example, a difference of 100 rating points leads to an expected win rate of 64\,\% for the higher rated player.
After each match, the winner gains points and the loser loses points, with the magnitude of the update depending on the difference between actual and expected outcomes.
Hence, this analysis complements the displacement error metric, by providing a relative measure of how convincing the modified scenarios appear to a human expert.
In addition, we also compute the model rank, which is defined as one plus the number of other models whose lower confidence interval bound is higher than the upper confidence interval bound of the current model.

The results show that in a blind comparison, scenarios created using GPT-4o \texttt{OTM} are almost indistinguishable from the human generated scenarios from interPlan.
At the same time, \texttt{OTM} with the two smaller models Gemini-1.5-Flash and Llama3.1-70B are significantly weaker in terms of Elo.
Interestingly, between \texttt{FC}, \texttt{tQA} and \texttt{vQA}, the human judges expressed a preference towards QA variants with \texttt{vQA} being almost as good as GPT-4o \texttt{OTM}.
This stands in contrast to the displacement error metrics, where \texttt{FC} performs better.

\begin{figure*}
    % placeholders
    \centering

    \begin{subfigure}{\textwidth}%
        \centering
    % Row for headers
    \begin{minipage}{0.33\textwidth}
        \centering
        interPlan~\cite{Hallgarten2024interPlan} (\textbf{GT})\\
        % \vspace{0.2em}
        \includegraphics[width=\textwidth]{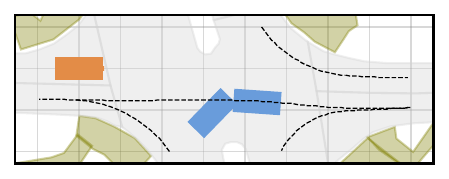}%
    \end{minipage}%
    \hfill
    \begin{minipage}{0.33\textwidth}
        \centering
        \textbf{Ours} (GPT \texttt{OTM})\\
        % \vspace{0.1em}
        \includegraphics[width=\textwidth]{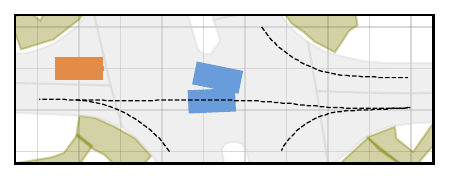}%
    \end{minipage}%
    \hfill
    \begin{minipage}{0.33\textwidth}
        \centering
        \textbf{Ours} (Gemini \texttt{OTM})\\
        % \vspace{0.2em}
        \includegraphics[width=\textwidth]{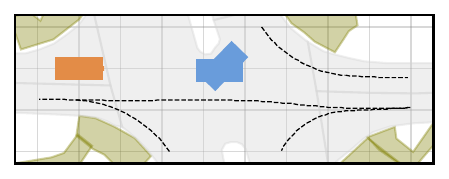}%
    \end{minipage}
        \vspace{-0.1cm}
        \caption{Generating an accident site in front of ego.}
        \label{fig:accident}
    \end{subfigure}
    \vspace{-0.2cm}
    
    \begin{subfigure}{\textwidth}%
    \centering
    % Row for headers
    \begin{minipage}{0.33\textwidth}
        \centering
        interPlan~\cite{Hallgarten2024interPlan} (\textbf{GT})\\
        % \vspace{0.2em}
        \includegraphics[width=\textwidth]{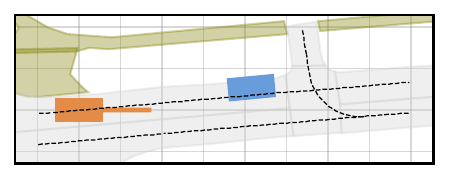}%
    \end{minipage}%
    \hfill
    \begin{minipage}{0.33\textwidth}
        \centering
        \textbf{Ours} (GPT \texttt{OTM})\\
        % \vspace{0.1em}
        \includegraphics[width=\textwidth]{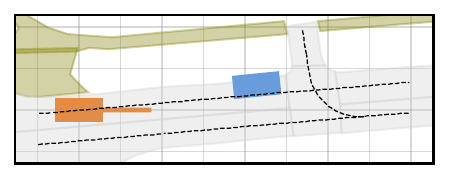}%
    \end{minipage}%
    \hfill
    \begin{minipage}{0.33\textwidth}
        \centering
        LCTGen~\cite{tan2023language}\\
        % \vspace{0.2em}
        \includegraphics[width=\textwidth]{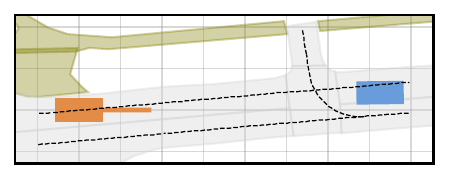}%
    \end{minipage}
        \vspace{-0.1cm}
        \caption{Adding a parked car in front of the intersection.}
        \label{fig:parked}
    \end{subfigure}
    \vspace{-0.2cm}

    \begin{subfigure}{\textwidth}%
    \centering
    % Row for headers
    \begin{minipage}{0.33\textwidth}
        \centering
        interPlan~\cite{Hallgarten2024interPlan} (\textbf{GT})\\
        % \vspace{0.2em}
        \includegraphics[width=\textwidth]{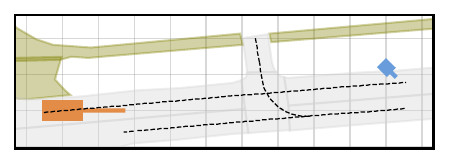}%
    \end{minipage}%
    \hfill
    \begin{minipage}{0.33\textwidth}
        \centering
        \textbf{Ours} (GPT \texttt{OTM})\\
        % \vspace{0.2em}
        \includegraphics[width=\textwidth]{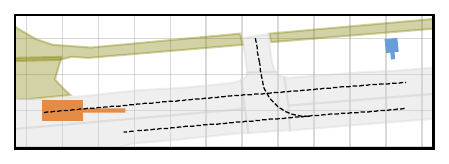}%
    \end{minipage}%
    \hfill
    \begin{minipage}{0.33\textwidth}
        \centering
        \textbf{Ours} (Gemini \texttt{FC})\\
        % \vspace{0.2em}
        \includegraphics[width=\textwidth]{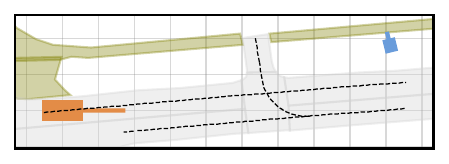}%
    \end{minipage}
        \vspace{-0.1cm}
        \caption{Adding a pedestrian crossing the road in front of ego.}
        \label{fig:pedestrian}
    \end{subfigure}
   
    \caption{Qualitative samples for three different scenario types. Ego vehicle in red, modified traffic agents in blue, drivable area in gray and walkways in olive. Grid represents 5\,m intervals. (\subref{fig:accident}): Gemini placed the accident vehicles at the correct distance, but with an unrealistically large overlap. (\subref{fig:parked}): LCTGen placed the vehicle behind the intersection and on the wrong lane. Despite moderate displacement error, this misses the intention of the user. (\subref{fig:pedestrian}): Gemini placed the pedestrian at the right distance, but facing away from the road.}
    \label{fig:samples}
    \vspace{-0.5cm}
\end{figure*}

In summary, we observe the general trend that GPT-4o performs well across the board, despite the simplicity of the \texttt{OTM} variant, but its commercial nature raises the questions of cost and reliance on a closed-source API.
However, with the advanced prompting strategies offered by our framework, we can leverage cheaper models and achieve similar performance.
Specifically, the \texttt{vQA} variant allows Gemini-1.5-Flash, a relatively cheap utility model, to close the gap to frontier models in terms of visual appearance.
For qualitative samples see \autoref{fig:samples}.

\begin{table}
    \centering
    \begin{tabular}{l>{\columncolor{llm_blue!20}}c r r r}
    \toprule
    \textbf{Model} & \textbf{Rank} & \textbf{Elo}~$\uparrow$ & \textbf{95\% CI} & \textbf{Votes}\\
    \midrule
    interPlan & 1 & 1042 & -9/+11 & 1960\\
    GPT-4o \texttt{OTM} & 1 & 1039 & -9/+11 & 1720\\
    Gemini-1.5-Flash \texttt{vQA} & 1 & 1025 & -12/+13 & 720\\
    Llama3.1-70B \texttt{tQA} & 3 & 1011 & -16/+15 & 600\\
    Gemini-1.5-Flash \texttt{tQA} & 3 & 1003 & -15/+15 & 600\\
    Gemini-1.5-Flash \texttt{FC} & 4 & 998 & -10/+9 & 1360\\
    Llama3.1-70B \texttt{FC} & 5 & 984 & -8/+10 & 1360\\
    Gemini-1.5-Flash \texttt{OTM} & 8 & 953 & -12/+12 & 1600\\
    Llama3.1-70B \texttt{OTM} & 8 & 941 & -13/+11 & 1600\\
    \bottomrule
    \end{tabular}
    \caption{Elo with 95\% Confidence Intervals.}
    \label{tab:elo}
    \vspace{-0.5cm}
\end{table}

\subsection{Benchmarking SotA Planning Methods}

Since the purpose of our framework is to create challenging scenarios for testing and verification of AD planners, we run our scenarios in a closed-loop simulation using the nuPlan framework.
We employ the PDM-Closed planner, which is the top-performing method in nuPlan.
This sampling-based planner first generates several candidate trajectories by rolling out multiple IDM-policies assuming constant-velocity predictions for all traffic agents.
These IDM-policies vary in their target speeds and lateral offset from the centerline.
Unlike the original method, which combines five target speeds and three lateral offsets ($-1\,m$, $0\,m$, $+1\,m$), we increase the number of proposals by sampling offsets up to $\pm 4\,m$ to allow the planner to deviate further from the centerline - a capability that is crucial for scenarios where the centerline is blocked, e.g., by a parked vehicle.
As in the original method, each proposal is evaluated for safety, progress and comfort and the best one is selected.
If no proposal is free of infractions, then the planner brakes and remains stationary, thus not making any progress.

In nuPlan, a planner's performance in a scenario is evaluated using a driving score that aggregates metrics based on safety, comfort, and progress~\cite{nuplan}.
Besides metrics for progress and comfort, the time-to-collision is computed.
These are compared to a threshold and aggregated into a weighted average, which is multiplied by penalties for drivable area-infractions and collisions.
Penalties are $1$ if no infraction occurs throughout the 15\,s of simulation and $0$ otherwise.

\autoref{tab:driving_score} shows the simulation score averaged across all 50 scenarios for the \texttt{OTM} variants of our framework.
For reference, we also included scores for interPlan and the Val14 test split of nuPlan.
We observe that the score on Val14 is saturated as it is close to a perfect score.
interPlan~\cite{Hallgarten2024interPlan} introduces difficult scenarios which leave room for improvement on the planner side.
However, it is limited to a few hand-crafted scenarios.
Our method can generate equally challenging scenarios in a semi-automatic setup.
We hope that this can fuel more research on more sophisticated planning methods.

\begin{table}
    \centering
    \begin{tabular}{lr}
    \toprule
    \textbf{Scenarios} & \textbf{Mean Driving Score} [\%] \\
    \midrule
    Val14 & 90.8 \\
    interPlan & 51.9 \\
    \midrule
    GPT-4o \texttt{OTM} & 49.6 \\
    Gemini-1.5-Flash \texttt{OTM} & 53.5 \\
    Llama3.1-70B \texttt{OTM} & 54.0 \\
    % GPT-4o \texttt{OTM} & 54.7 \\
    % Gemini-1.5-Flash \texttt{FC} & 57.8 \\
    % Llama3.1-70B \texttt{FC} & 55.7 \\
    \bottomrule
    \end{tabular}
    \caption{Mean driving score of PDM-Closed.}
    \label{tab:driving_score}
    \vspace{-0.5cm}
\end{table}

\section{Conclusion}
In this paper, we introduced a framework for modification and augmentation of traffic scenarios using natural language.
By introducing a formalized, text-based scenario description format, we are able to leverage an LLM-based agentic framework.
Using frontier models, our framework achieved comparable output quality as human generated scenarios in a blind, side-by-side comparison with human domain experts as judges.
Additionally, our scenarios proved equally challenging as the human generated scenarios to PDM-Closed, a sotA planner.
One limitation is the dependence on commercial frontier LLMs, which are only accessible through commercial APIs.
By employing advanced prompting techniques like function calling or QA agents, we were able to narrow the performance gap between frontier and utility/open-weight models.
This represents a significant opportunity to reduce cost and reliance on closed-source APIs.
We are confident that further improvement in prompting techniques and tool use, in combination with the rapid improvement of open-weight LLMs, will close this gap completely.
The code for this paper will be made available.

\bibliographystyle{IEEEtran}
\bibliography{bibliography}
% \appendix
% \subfile{07_appendix.tex}

\end{document}